\documentstyle[twocolumn,fleqn,12pt]{article}
\input{psfig}
\input{psfig}

\begin{document}

\title{\vspace{-2em}
{\small{
Appears in the {\it Proceedings of the International Workshop on
Sharable Natural Language Resources}, \\
\vspace{-1em}
Nara, Japan, August~1994, pp.~123-130}}
\vspace{.5em}
\\A Freely Available Syntactic Lexicon for English}
\author{
Dania Egedi and Patrick Martin\thanks{Currently at SRA, Arlington, VA, 22201 USA; martinp@sra.com} \\
Institute for Research in Cognitive Science \\
University of Pennsylvania \\
Philadelphia, PA 19104-6228, USA \\ \\
\{egedi,martin\}@unagi.cis.upenn.edu
}
\date{}
\maketitle

\begin{tabular*}{6.25in}[c]{ccc}
\hspace*{2.6in}&{\bf Abstract} &\hspace*{2.6in} \\
\end{tabular*} \\

\begin{tabular*}{6.25in}[c]{l}
This paper presents a syntactic lexicon for English that was originally derived
from the \\Oxford Advanced Learner's Dictionary and the Oxford Dictionary of
Current Idiomatic \\English, and then modified and augmented by hand.  There
are more than 37,000 syn- \\tactic entries from all 8 parts of speech.  An
X-windows based tool is available for main- \\taining the lexicon and performing
searches.  C and Lisp hooks are also available so that \\the lexicon can be
easily utilized by parsers and other programs.
\end{tabular*} \\


\bibliographystyle{plain}

\section{Introduction}

One of the central needs of any wide-coverage parser is a large lexicon that
contains the syntactic information for various lexical items.  The creation of
such a lexicon has traditionally been a very large and daunting task and most
universities have shied away from it, leaving the creation of wide-coverage
parsers to commercial institutions that could afford the time and personnel to
devote to the creation of such a lexicon.  The release of several
machine-readable dictionaries (MRDs) into the public domain has opened new
possibilities to grammar developers at research institutions, but the task
did not become trivial.  The problem of creating large scale lexicons changed
from the tiresome, painstaking task of trying to develop individual word lists
for various syntactic phenomena to the task of `simply' extracting the
information from the on-line dictionaries.  This, however, has not turned out
to be as simple or straight-forward as researchers may have hoped.  Machine
readable dictionaries present numerous problems in terms of errors and in-\\

\vspace*{1.7in}
\noindent
consistencies in the various components of the lexical entries, making
extraction quite difficult.  Many researchers abandon the extraction process
altogether because it consumes too many scarce resources.

Although a number of researchers have extracted information out of the
various dictionaries available, the resulting lexicons have not, in general, been made
freely available to the NLP research community.  In at least some cases
(\cite{carroll89},
\cite{guthrie93}) this is due to licensing restrictions on the source 
dictionaries.  In response to the related problems of duplication of effort and
non-availability of needed lexicons, there are currently several on-going
projects to create syntactic lexicons and make them generally available.

\begin{itemize}

\item  The Proteus Project at New York University is developing the Comlex 
Syntactic Dictionary from scratch for release as one of the lexical resources 
in COMLEX (available through the Linguistic Data Consortium) \cite{macleod94}.

\item The IITLEX project at Illinois Institute of Technology has an
on-going project to extract and release the information in the Collins English
Dictionary, along with information from various other word lists that will
include both syntactic and semantic information.  That system is still under
development, however, and currently uses an expensive relational database
package, a drawback which they plan to correct. \cite{conlon94}

\end{itemize}

The syntactic lexicon described here contains approximately 37,000 entries
extracted from the {\it Oxford Advanced Learner's Dictionary of Current
English} \cite{oald74} and the {\it Oxford Dictionary for Current Idiomatic
English} \cite{cie75}.  It is available via FTP in both an ASCII and a database
format.  The database format uses a UNIX hash table facility \cite{sy91} that
is freely distributed, and comes with an X-windows based interface for
modifying the database and doing searches.  C and Lisp hooks to allow other
programs to use the database are also included.

\section{Syntactic Lexicon}

The syntactic lexicon has entries for 8 part-of-speech categories: Adjective,
Adverb, Complementizer, Conjunction, Determiner, Noun, Preposition, and Verb.
Each entry consists of the following required and optional fields:

\begin{itemize}

\item {\sc index} field (required) -- the uninflected form under which
the lexical item is compiled in the database; 

\item {\sc entry} field (required) -- contains all of the lexical items 
associated with the {\sc index}\footnote{For example, a verb particle
construction would be {\sc index}ed under the verb, but would contain both the
verb and the verb particle in the {\sc entry} field.};

\item {\sc pos} field (required) -- gives the part-of-speech for the lexical 
item(s) in the {\sc entry} field;

\item {\sc frame} field (required) -- contains the syntactic information about 
that entry;

\item {\sc fs} field (optional) -- the Feature Structure field may provide
additional information about the {\sc frame} field.  

\item {\sc ex} field (optional) --  may be used for any number of example 
sentences.

\end{itemize}

Note that lexical items may have more than one entry in the database (e.g. {\it
have}) and that they may select the same {\sc frame} field more than once,
using the {\sc fs} to capture lexical idiosyncrasies (e.g. {\it map}).
Table~\ref{syn-entries} shows selected entries from the database.

\begin{table}[ht]
\begin{tabular}{ll}

INDEX:& have \\
ENTRY:& have \\ 
POS:& Verb \\ 
FRAME:& Auxiliary\_Verb \\
FS:& Goes\_on\_Infinitive  \\ 
EX:& John has to go to the store. \\
\\
INDEX:&have\\
ENTRY:& have \\
POS:& V \\
FRAME: &Transitive\_Verb \\
FS:& Non-Ergative \\
EX:& John has a problem. \\
\\
INDEX:&map \\
ENTRY:&map out \\
POS:&Verb Verb\_Particle \\
FRAME: & Transitive\_Verb\_Particle \\
\\
INDEX:& map \\
ENTRY:& map \\
POS:& Noun \\
FRAME:& Base\_Noun \\
&Noun\_Determiner\_required\\
&Noun\_Modifier\\
FS: &wh$-$, reflexive$-$\\
\\
INDEX:&map \\
ENTRY:&map \\
POS:&Noun \\
FRAME:&Noun\_Determiner\_not\_required\\
FS:&wh$-$, reflexive$-$, plural

\end{tabular}	
\caption{Selected Syntactic Database Entries}
\label{syn-entries}
\end{table}

Because the syntactic database is part of the XTAG project \cite{xtag-notes94},
a on-going project to develop a wide-coverage parser for English (see Section
\ref{related-work}), some entries in the syntactic lexicon reflect specific 
XTAG analyses.  In fact, the graphical interface for the syntactic lexicon
(described in Section~\ref{interface}) can run in two modes - {\bf xtag} and
{\bf verbose}.  Tables \ref{syn-entries}, \ref{verb-entries}, and
\ref{aux-verbs} were all generated in {\bf verbose} mode.

The vast majority of lexical items in the database fall into just 3 categories
- Adjectives, Nouns, and Verbs.  These three categories plus Adverbs are
presented in more detail in the following subsections.

\subsection{Adjectives}

There are 3,303 lexical adjectives in the database, of which 80 are `Proper
Name' adjectives, such as {\it Chinese} and {\it American}.  Adjectives have 5
frames that they can select, which are listed below.  Possible values for the
{\sc fs} field are {\bf wh$-$} and {\bf wh$+$}.

\begin{itemize}

\item {\bf Base adjective}:  All adjectives.

\item {\bf Modifying adjective}: Adjectives that can occur in direct
modification contexts.  Ex. {\it the {\bf Chinese} man.}

\item {\bf Predicative adjective}: Adjectives that can occur as the complement
of a predicative verb. Ex. {\it John was {\bf happy}.}

\item {\bf Predicative adjective w/ sentential complement}: Adjectives that can
occur as the complement of a predicative verb and that take a sentential
complement.  Ex. {\it John was {\bf happy} that Mary left Bill.}

\item {\bf Predicative adjective w/ sentential subject}:  Adjectives that can
occur as the complement of a predicative verb and that take a sentential
subject.  Ex. {\it That John loves Mary is {\bf great}!}

\end{itemize}

\subsection{Nouns}

Nouns are by far the largest category in the syntactic database, accounting for
well over 50\% of the entries.  Proper nouns and pronouns both have the
part-of-speech Noun.  Proper names, such as {\it Danielle} and {\it Nicholas}
are not well-represented in the database, but geographic names, particularly
places in England, generally are\footnote{This reflects the origin of the
dictionary from which the lexicon was originally extracted.}.  The frames for
nouns are similar in many ways to the frames for adjectives, since nouns can
modify other nouns and occur in predicative sentences.  Other frames provide
information about the use of the noun with determiners when forming noun
phrases.  The frames for noun are presented below:

\begin{itemize}

\item {\bf Base noun}:  All nouns.

\item {\bf Noun Phrase with Determiner}: Nouns that can take a determiner when
forming a noun phrase.  Ex. {\it a {\bf man}}; *{\it a {\bf jealousy}}

\item {\bf Noun Phrase without Determiner}: Nouns that can appear without a
determiner when forming a noun phrase.  Ex. {\it {\bf envy}}; *{\it {\bf
plant}}

\item {\bf Modifying noun}: Nouns that can modify other nouns.  Note that not
all nouns can modify other nouns.  Proper nouns in general cannot modify other
nouns, and specific lexical items may be restricted as well. Ex. {\it
{\bf basketball} game}; *{\it {\bf John} car}

\item {\bf Noun with sentential complement}: Nouns that take sentential
complements.  Ex. {\it the {\bf fact} that Mary loves John...}

\item {\bf Predicative noun}: Nouns that can occur as the complement
of a predicative verb. Ex. {\it John was a {\bf man}.}

\item {\bf Predicative noun w/ sentential subject}:  Nouns that can
occur as the complement of a predicative verb and that take a sentential
subject.  Ex. {\it That John loves Mary is a {\bf crime}.}

\end{itemize}

Because this lexicon is used in the XTAG system, the lexicon often indicates
precise syntactic behavior, rather than simply placing a general label on a
lexical item.  For the class of nouns, this is seen in the specification of
nouns with respect to their co-occurrence with determiners.  Instead of
assigning a general label as as `common noun' or `mass noun', the noun frames
explicitly indicate whether certain forms of the noun can appear with or
without a determiner.  However, since the syntactic database is indexed on root
forms only, the morphology of the lexical item is not available.  Instead, the
{\sc FS} field is used to indicate any restrictions on a particular use of a
lexical item.  For example, in Table~\ref{syn-entries}, the noun {\it map}
occurs twice.  The first time that it appears, it selects the {\bf
Noun\_Determiner\_required} frame.  The feature structures associated with it
indicates only that the noun is not a wh-word, and that it is not reflexive.
No restrictions are made with respect to its morphology.  In contrast, the
second entry, which selects the {\bf Noun\_Determiner\_not\_required} has {\bf
plural} as part of its {\sc FS}.  This indicates that the noun for this frame
is restricted to its plural form.  Hence {\it map} can only occur with a
determiner, but {\it maps} is free to occur both with or without one.  Nouns
that belong to the class of so-called `mass nouns' would not have the {\bf
plural} restriction on the entry that selects the {\bf
Noun\_Determiner\_not\_required} frame, thereby indicating that the singular
form is also allowed to occur without a determiner.

\subsection{Verbs}

Verbs, with their varied subcategorization frames, are perhaps the most
interesting lexical items in a syntactic lexicon.  There are over 8100
verbs (not including auxiliary verbs) that make up almost 9000 entries
in the database.  There are 19 different frames that the verbs can select,
including transitive, intransitive, sentential complement, sentential subject,
verb particle constructions (transitive and intransitive), double objects with
shifting, double objects without shifting, and light verb constructions.

As with the nouns, the {\sc FS} field is used to provide a more concise format
for specifying the frames for each lexical item.  For the verbs, the {\sc FS}
field is used to specify the difference between ergative and non-ergative
transitive verbs, as can be seen in the {\it have} entry in
Table~\ref{syn-entries}, and is also used heavily for further differentiating
the frames for verbs that take sentential complements.  There are two frames
for sentential complements - {\bf Sentential\_Complement} and {\bf
NP\_and\_Sentential\_Complement}.  Either of these can occur with the feature
structures {\bf Infinitive\_Complement}, {\bf Indicative\_Complement}, or {\bf
Predicative\_Comp\-le\-ment}.  This reduces the number of values for {\sc
FRAME} that are necessary to cover all of the possible lexical environments,
and also allows for easier searches across categories.  To find all the verbs
that take infinitive complements, one can simply search on the {\bf
Infinitive\_Complement} feature structure, rather than having to specify each
frame that could fill this role.  Table~\ref{verb-entries} shows some values
for various verbs that take sentential complements.

\begin{table}[ht]
\begin{tabular}{ll}

INDEX: &want \\
ENTRY: &want \\ 
POS: &Verb \\ 
FRAME: &Sentential\_Complement \\
FS: &Infinitive\_Complement   \\ 
EX: &Dan wants to finish this paper. \\
\\
INDEX: & want \\
ENTRY:& want \\ 
POS:& Verb \\ 
FRAME:& NP\_and\_Sentential Complement \\
FS:& Infinitive\_Complement\\ 
EX:& Dan wants Al to finish this paper. \\
\\
INDEX: &think \\
ENTRY: &think \\ 
POS:& Verb\\ 
FRAME:& Sentential\_Complement \\
FS:& Indicative\_Complement\\ 
EX:& Dan thought that the paper was done. \\
\\
INDEX:& think \\
ENTRY:& think \\ 
POS:& Verb\\ 
FRAME:& Sentential\_Complement \\
FS:& Infinitive\_Complement\\ 
EX:& Doug thought to clean the kitchen.\\
\\
INDEX:& think \\
ENTRY:& think \\ 
POS:& Verb\\ 
FRAME:& Sentential\_Complement \\
FS:& Predicative\_Complement\\ 
EX:& Dan thought Carl a jerk.

\end{tabular}	
\caption{Verbs with Sentential Complements}
\label{verb-entries}
\end{table}

\subsubsection{Auxiliary verbs}

The lexical entries for auxiliary verbs are very closely tied to the XTAG
analysis, which orders the auxiliary verbs based on their morphological
forms.  Each entry in the lexicon is restricted via the {\sc FS} field to only
a certain form of the auxiliary verb ({\bf present}, {\bf past}, {\bf
ppart}, etc), which also indicates what other forms that it can go
on\footnote{For a more detailed description of this and other XTAG analyses,
please see the XTAG Technical Report \cite{xtag-tech}.}.  Table
\ref{aux-verbs} shows the entries for the auxiliary verbs for the sentence {\it
John should have been waiting.}

\begin{table}[ht]
\begin{tabular}{ll}

INDEX: &should \\
ENTRY: &should \\ 
POS:& Verb \\ 
FRAME:& Auxiliary\_Verb\\
FS:& Indicative, Present, Goes\_on\_Base  \\ 
\\
INDEX:& have \\
ENTRY:& have \\ 
POS:& Verb \\ 
FRAME:& Auxiliary\_Verb\\
FS:& Base, Goes\_on\_Past\_Participle  \\ 
\\
INDEX:& be \\
ENTRY:& be\\ 
POS:& Verb \\ 
FRAME:& Auxiliary\_Verb\\
FS:& Past\_Participle, Goes\_on\_Gerund

\end{tabular}	
\caption{Example Auxiliary Verb Entries}
\label{aux-verbs}
\end{table}

\subsection{Adverbs}

A syntactic lexicon for adverbs is particularly useful because adverbs are so
idiosyncratic as to where they can occur in a sentence.  Although there are
only 169 adverbs in the syntactic lexicon, but there are 15 different {\sc
frame} values that they can select.  These include basic adverb, pre and post
verb phrases, pre and post sentences, pre and post adjective, pre-adverb,
pre-preposition, pre-noun, etc.  Table \ref{adv-entries} shows some selected
adverb entries.

\begin{table}[ht]
\begin{tabular}{ll}

INDEX:& ahead \\
ENTRY:& ahead \\
POS:&	Adverb \\
FRAME:& Base\_Adverb \\
& Post-VP \\
& Pre-PP \\
\\
INDEX:& essentially \\
ENTRY:& essentially \\
POS:& Adverb \\
TREES:& Base\_Adverb \\
& Pre-VP \\
& Pre-S \\
& Post-S \\
\\
INDEX:& even \\
ENTRY:& even \\
POS:& Adverb \\
FRAME:& Base\_Adverb \\
& Pre-VP \\
& Pre-Adj \\
& Pre-Noun \\
& Pre-PP \\
\\
INDEX:& very \\
ENTRY:& very \\
POS:& Adverb \\
FRAME:& Base\_Adverb \\
& Pre-Adj \\
& Pre-Adv

\end{tabular}	
\caption{Some Adverb Lexical Entries}
\label{adv-entries}
\end{table}

\section{File Formats}

The information in the syntactic database is available both in an ASCII 'flat'
file, and a hashed database format.  The ASCII file contains one entry per
line, and each field is clearly marked.  This format is easily usable by
various UNIX$^{tm}$ utilities such as {\it grep} and {\it awk}, and it can be
easily parsed by custom programs.  

The hashed database format is very useful for programs that need quick access
to the information in the database.  Each entry is indexed under the {\sc
index} key, and a single call to the database for a particular index returns
all of the entries that share that index.  This makes it particular useful for
parsers.  The database uses an encoding scheme for the {\sc pos}, {\sc frame},
and {\sc FS} fields, which condenses the space required for the database and
shortens the search time for non-index fields.  All of the entries for a given
lexical index can be retrieved in 1.6 msecs, on average.

\section{Interface}
\label{interface}

Although the format of the flat file is excellent for various file utilities
programs, and the database format works well for retrieving entries quickly,
neither is particularly well-suited for human readability.  The X-windows
interface{\footnote{The interface uses the MIT Athena Toolkit, which is
distributed with the standard MIT X release.} for the syntactic database allows
users to easily look at the database.  Searching is available not only on the
{\sc index} under which the lexical item is stored, but also on all other
fields, with the exception of the {\sc ex} field.  Searches may also be done on
combinations\footnote{We hope to add expand this in the future to include full
regular expression searches.} of fields.  For instance,
one could search on {\sc POS}~=~{\bf Noun} and {\sc FS}~=~{\bf wh$+$} to find
the set of all wh+ nouns ({\it what}, {\it who}, {\it whom}, {\it which}, {\it
when}).  Figure \ref{interface-fig} shows the interface after a search has been
done on the index {\it need}.  All of the entries with that index are listed in
a scroll window, which can be browsed through using the {\bf Next} and {\bf
Previous} buttons, or specific entries can be clicked on, and the entire record
will show in the upper window.  The results of searches can be saved to a file
to create smaller `custom' lexicons.  In addition to searching the database,
users can also easily add, delete and modify individual entries, tailoring the
syntactic database to fit their needs.  Users may also delete all entries found
in a given search, and we hope to add the capacity to modify a entire set of
entries in the future.

\begin{figure}[htb]
\centering
{\psfig{figure=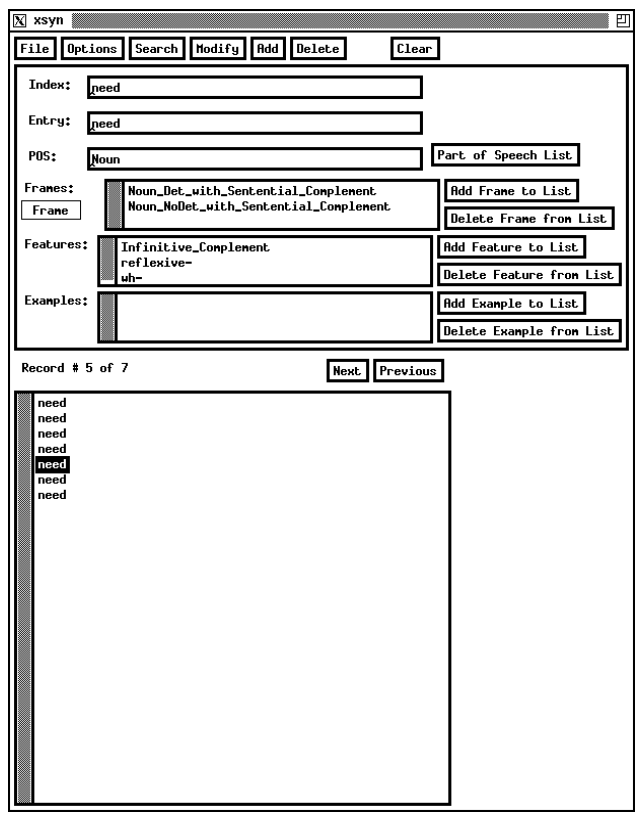,height=4.2in}}
\caption{Result of a search on the index {\it need}}
\label{interface-fig}
\end{figure}

\section{Statistics}

Statistics were gathered on the coverage of the syntactic lexicon on the IBM,
ATIS, WSJ, and Brown corpora.  These corpora were chosen because they have been
tagged and hand corrected by the TreeBank project \cite{santorini90}.  The data
in Table \ref{corpora-stats} show the coverage of the lexicon on various
corpora.  A lexical item/part-of-speech pair is counted as a {\bf hit} if the
lexical item is in the syntactic lexicon with the indicated tag.  No attempt
was made to determine if the lexicon had the correct frame needed to parse the
sentence.

\begin{table}[htb]
\begin{center}
\begin{tabular}{|l|r|r|c|} \hline
& Number & Total \# & Percent\\
Corpus & of Hits & of Words & Hit\\ \hline

WSJ &  1974528 &  2462557 & 80.18\% \\\hline

Brown & 799904 & 991008 & 80.72\% \\\hline

IBM  & 60944 & 68800 & 88.58\%\\ \hline

ATIS &  10156 &  13791 & 73.64\%\\ \hline
\end{tabular}
\caption{Percentage of Hits for various corpora}
\label{corpora-stats}
\end{center}
\end{table}

\begin{table*}[ht]
\begin{center}
\begin{tabular}{|l|r|c|c|c|c|c|} \hline
& Number of & Percent & Percent & Percent & Percent & Percent\\
Corpus &  Non-hits & Proper N & Nouns & Adj & Adv & Verbs\\ \hline

WSJ & 488029  & 43.8\%  & 30.7\% & 13.8\% & 5.7\% & 1.3\% \\ \hline

Brown & 191104 & 26.2\% & 40.6\% & 14.8\% & 7.4\% & 1.8\% \\ \hline

IBM  & 7856 & 17.1\% & 56.9\% & 11.3\% & 2.8\% & 2.5\% \\ \hline

ATIS & 3635  & 67.4\%  & 14.0\% & 1.6\% & 0.6\% & 2.4\% \\ \hline
\end{tabular}
\caption{Percentage of missing words for various Parts of Speech}
\label{missing-stats}
\end{center}
\end{table*}

Because the syntactic lexicon contains only the root form of lexical entries,
the inflected form was first looked up in the morphology database \cite{karp92}
to retrieve the root form, and then that was used for the syntactic lexicon.
Items that were not found in the morphological database were counted against
the syntactic lexicon, as the morphology database is a superset of the
syntactic database\footnote{Because these databases are being used in an actual
parser, an attempt was made some time ago to make ensure that all words in the
syntactic lexicon appear in the morphological database.  Although the databases
may have diverged slightly since then, it should not be statistically
significant.}.  The statistics in Table \ref{corpora-stats} are over all word
occurrences in the corpora\footnote{Numbers and the genitive marker ('s) were
taken out before the statistics were compiled.}, so words that occur frequently
are given more weight.

Not surprisingly, nouns and proper nouns\footnote{Although we do not
distinguish nouns and proper nouns in the syntactic lexicon, the TreeBank tags
do make this distinction, and it seemed useful to continue this distinction for
this part of the analysis.} comprise the largest category of words missed,
followed by adjective, adverbs, and verbs.  Table \ref{missing-stats} shows the
percentage of each of these categories in the list of items not found.  Again,
this is a percentage of word occurrences in the corpora.

As Table \ref{missing-stats} indicates, the majority of the missing items are
either nouns or proper nouns (66.8\% - 81.4\%).  This is not surprising, nor
particularly distressing, as nouns tend to be the easiest items to `guess'
information about.  Verbs, which tend to be the hardest, are reasonably
well-covered in this lexicon.  The number of adjectives not covered, however,
seems fairly high, and we plan to add a number of those missing to the
syntactic lexicon.

\section{Future Work}

The lexicon in its present form does not provide a mechanism to specify
preferences of lexical items for certain syntactic structures.  As part of
future enhancements to the lexicon we hope to associate probabilities with each
entry.  The probabilities will reflect the affinity of the lexical item for the
syntactic structure associated with that entry.  These probabilities will be
computed from parsed corpora.

It has been observed quite conclusively in recent work in lexicography that
certain combinations of words co-occur more often than would be expected if
they corresponded to arbitrary usages of the individual words.  Collocational
information has been shown to be of immense use in pruning the search space for
a parser. We hope to eventually extract collocational
information from the corpora and make it a part of the syntactic lexicon.

\section{Related Work}
\label{related-work}

The syntactic lexicon was developed as part of the XTAG project
\cite{xtag-notes94} at the University of Pennsylvania under the direction of 
Dr.  Aravind Joshi.  The XTAG system is a wide-coverage parser and grammar for
English based on the Tree Adjoining Grammar (TAG) formalism \cite{joshi75}.
The English grammar consists of 3 sections - a morphology database, a syntactic
database, and a tree grammar.  Together with a parser and an X-windows
interface, they comprise the XTAG system.  Both the morphology \cite{karp92}
and syntactic databases are available separately.  The entire XTAG system is
also freely available to the NLP research community.  Information about the
entire XTAG system and FTP instructions may be obtained by writing
xtag-request@linc.cis.upenn.edu.

\section{Computer Platform}

The syntactic lexicon and accompanying interface were developed on the Sun
SPARC station series, as were the other tools mentioned in
Section~\ref{related-work}.  All of the XTAG tools, including the syntactic
lexicon and interface, are freely available without limitation through
anonymous FTP to {\bf ftp.cis.upenn.edu}.  The syntactic lexicon and
accompanying programs together require about 9MB of space (for both the ASCII
and DB versions of the lexicon).  Please send mail to
lex-request@linc.cis.upenn.edu for current FTP instructions or for more
information.

\newpage
\bibliographystyle{aaai-named}

\end{document}